\documentclass[sigconf]{acmart}
\usepackage{xcolor}
\usepackage{hyperref}
\renewcommand\footnotetextcopyrightpermission[1]{} 

\AtBeginDocument{%
  }

\setcopyright{none}
\acmConference[]{}{}{}
\acmBooktitle{}
\acmPrice{}
\acmDOI{}
\acmISBN{}

\begin{document}

\title{An Empirical Study of Causal Relation Extraction Transfer: Design and
Data}

\author{%
  Sydney Anuyah\href{https://orcid.org/0000-0003-1929-841X}, 
  Jack Vanschaik\href{https://orcid.org/0000-0003-3663-207X}, 
  Palak Jain, 
  Sawyer Lehman, 
  Sunandan Chakraborty\href{https://orcid.org/0000-0002-3331-6082}.}%
  
\affiliation{%
  \institution{Indiana University, Indianapolis}
  \state{Indiana}
  \country{USA}}
  
\email{ sanuyah@iu.edu, jainpalak3286@gmail.com, sblehman@iu.edu, sunchak@iu.edu}

\renewcommand{\shortauthors}{Sydney et al.}

\begin{abstract}
We conduct an empirical analysis of neural network architectures and data transfer strategies for causal relation extraction. By conducting experiments with various contextual embedding layers and architectural components, we show that a relatively straightforward BioBERT-BiGRU relation extraction model generalizes better than other architectures across varying web-based sources and annotation strategies. Furthermore, we introduce a metric for evaluating transfer performance, $F1_{phrase}$ that emphasizes noun phrase localization rather than directly matching target tags. Using this metric, we can conduct data transfer experiments, ultimately revealing that augmentation with data with varying domains and annotation styles can improve performance. Data augmentation is especially beneficial when an adequate proportion of implicitly and explicitly causal sentences are included.
\end{abstract}



\maketitle

\section{Introduction}
Causality is essential for understanding the relationships between variables, enabling us to identify actual cause-and-effect dynamics rather than mere correlations. This understanding is crucial in fields such as medicine~\cite{bollegala2018causality}, where identifying causal factors can lead to effective treatments, and finance~\cite{mariko2021financial}, where it informs decisions with wide-reaching impacts. In technology and AI, causality improves the reliability and interpretability of predictive models. Establishing causality leads to better outcomes across various domains, often requiring the extraction of causal relationships from free-text sources like scholarly articles and news reports, which is considered a key task in the field of Natural Language Processing (NLP).

Although the extraction of causal relationships may have direct applications to many downstream tasks \cite{zhou2024causalkgpt}, a major goal of causal relation extraction is to construct knowledge graphs that can facilitate downstream uses. ConceptNet \cite{speer2017conceptnet} and CauseNet \cite{heindorf2020causenet} are examples of this; however, the rule-based extraction methods used for these datasets limit the type of causal relations present in the knowledge graph. CauseNet, for example, identified causal sentences in Wikipedia and ClueWeb12 with syntactic templates then used a neural sequence tagger to label causes and effects in the identified sentences. Though this method was able to create the largest available causal knowledge base, it could be missing many implicit causal relations. In this sentence from CauseNet, ``Insomnia is often \underline{caused} by fear, stress, and anxiety." the relationship is made explicitly clear by the highlighted marker ``caused". Other markers (e.g., ``leads to", ``associates", ``trigger", etc.) comprise the source sentences underlying CauseNet. By contrast, take ``The clock struck twelve with a loud chime that made me jump." from \cite{uzzaman2013semeval}. Causal semantics are present, but the lack of a clear marker means such sentences would have been excluded from CauseNet. This under-representation of implicit sentences could bias knowledge graphs and knowledge bases when rule-based extraction is used on a large scale. 

We have yet to see a large-scale causal knowledge graph extracted using both implicit and explicit source sentences. This paper addresses the root cause of that problem; causal relation extraction models are trained on disparate benchmark datasets that vary significantly in lexical composition and annotation style. Recent efforts have employed increasingly complicated cause/effect/relation annotation schemes \cite{li2021causality}, which produce over-specified models that are not well suited for open-domain extraction. To address this, our analysis examines transfer across six unique causal relation datasets that span varying domains, annotation styles, and implicit/explicit causality makeups. An empirical analysis of architectural components including choice of embedding, recurrent unit, and usage of conditional random fields (CRF) is conducted. In particular, we see transformer-based contextual word embeddings \cite{devlin2018bert} with specialized pre-training datasets, such as BioBERT \cite{lee2020biobert} and RoBERTa \cite{liu2019roberta}, work best across all six target datasets. Experiments further reveal that across domains, common architectures for relation extraction, BiLSTM, and Conditional Random Field (CRF) \cite{li2021causality} are perhaps unnecessary for an open domain causal relation extraction model as these components barely improve performance at the expense of speed and generalizability over a simpler BiGRU. Our current focus is on extracting intra-sentence causal relations, as they are easier to identify and evaluate using existing benchmarks, providing a foundation for more complex tasks. While newer models like Mistral 7b, Llama, and GPT may offer better performance, this work is limited to BERT and Flair embeddings due to their balance of computational efficiency and robust performance in domain-specific tasks. These embeddings are also well-studied and integrated into existing NLP pipelines, making them a practical choice for understanding causality.

The introduction of our $F1_{phrase}$ metric allows for a fair comparison of transfer performance by recognizing the target of causal relation extraction as the cause/effect phrase subtrees of the dependency tree. The application of this metric reveals that many target datasets can benefit from data augmentation, regardless of domain and annotation styles. Using $F1_{phrase}$ and the best transfer models, we further see that improved transfer performance relates to implicit/explicit sentence makeup of a dataset and noun phrase coverage training size, rather than training data size. Causal relation extraction from the web sources is a challenging task, due to noise in the open domain and lack of consistency across viable training datasets. Through the lens of phrasal evaluation, our results imply that models like BioBERT-BiGRU are particularly well suited to causal relation extraction in this setting since they can best incorporate disparate data domains and annotation styles. Large causal knowledge bases have relied on rule-based methods for localizing causal relations \cite{heindorf2020causenet}, creating a bias for sentences with explicit expressions of causality. This work shows that BioBERT-BiGRU, trained on a collection of disparate data and evaluated with $F1_{phrase}$ offers a path towards richer extraction of causal knowledge from noisy web-based sources. Our work contributes:

\begin{enumerate}
    \item{An empirical evaluation of causal relation extraction models across a combination of datasets showed the high performance of specially pre-trained BERT embeddings in the transfer setting, as well as the unnecessariness of Bi-LSTM and CRF layers. }
    \item{The $F1_{phrase}$ metric which facilitates the empirical evaluation analysis.} 
    \item{Furthermore, we conduct additional qualitative analysis to determine what data characteristics contribute to transfer performance. We show that the proportion of implicit causal sentences in the data improves transfer, more so than merely increasing dataset size.}  Our code files and supplementary materials will be made available after the review process. 
\end{enumerate}

\section{Related Work}
An early example of a machine learning application for causal extraction is \citet{blanco2008causal}, modified in \cite{zhang2023storytree} where Bagging and Decision Trees were used to classify cause-and-effect information. A study by \citet{egami2018make} provides a conceptual framework for text-based causal inferences, generalizing a codebook function to fit several models and linking higher-dimensional text to lower-dimensional representations for making inferences and estimating treatment effects. While machine learning methods are tried-and-true, they cannot learn incrementally in real time and require supervision, limiting their usefulness in real-world applications.

Deep learning applications have been used in recent advances in causal extraction. For instance, \cite{roemmele2018encoder} used the Choice of Plausible Alternatives (COPA) framework to predict causally related events in stories. CausalTriad a minimally supervised method for determining causality that relies on discourse connectives and focused distributional similarity techniques was created by \cite{zhao2018causaltriad}. Transferred embedding has been employed in more recent research, such as the BiLSTM-CRF model developed by \cite{li2021causality}. Tests revealed that models with transferred embeddings performed better than those without. These models included Flair, ELMo, and BERT. As demonstrated by the BERT architecture utilized by \cite{khetan2022causal}, transformer-based architectures have also excelled in causal extraction. Training on data other than the target data provided results that were equivalent to or better than training and testing on the target data. We distinguish our work by using transferred embeddings across many architectures and testing a larger number of datasets. Our study also evaluates general-purpose architectures without assuming domain-specific information.

Not much work has investigated transferable causal relation extraction models specifically. \citet{moghimifar2020domain} combined a graph convolutional network for encoding dependency trees (GCE) with an adversarial domain adaptive module (ACE) for cross-domain causality classification and relation extraction. This approach is a step towards general-purpose, large-scale causal relation extraction, showing increased transfer performance across domains. However, it requires the model to explicitly learn different domains, which might not always be available in open-domain causal relation extraction. 
\begin{table*}
  \caption{Macro-averaged of F1 Scores for 10 iterations of Embedding Choice Experiments}
  \label{embedding}
  \centering
  \begin{tabular}{ccccccc}
    \toprule
    \textbf{Dataset} & \textbf{BERTBase} & \textbf{BioBERT} & \textbf{RoBERTa} & \textbf{FlairMulti} & \textbf{FlairNews} & \textbf{FlairFineTuned} \\
    \midrule
    \textbf{CausalTimeBank}    & \textbf{0.636} & 0.590 & 0.600 & 0.567 & 0.543 & 0.573 \\
    
    \textbf{CauseNet-cause}    & 0.847 & \textbf{0.848} & 0.841 & 0.843 & 0.805 & 0.826 \\
    \textbf{CauseNet-noncause} & 0.833 & \textbf{0.835} & 0.828 & 0.826 & 0.797 & 0.814 \\
    \textbf{FinCausal2020}     & \textbf{0.598} & \textbf{0.596} & 0.580 & 0.558 & 0.561 & 0.579 \\
    \textbf{MedCaus}           & 0.773 & \textbf{0.777} & \textbf{0.778} & 0.753 & 0.752 & 0.756 \\
    \textbf{SemEval 2018}      & 0.853 & 0.850 & \textbf{0.861} & 0.820 & 0.783 & 0.822 \\
    \bottomrule
  \end{tabular}
\end{table*}

\begin{table*}
  \caption{Macro-averaged Test F1 Scores for 10 iterations of RNN Experiments and Hypothesis Testing Results}
  \label{recurrent}
  \centering
  \begin{tabular}{ccccccc}
    \toprule
    \textbf{Dataset} & \textbf{BaseGRU} & \textbf{BaseLSTM} & \textbf{BioBERTGRU} & \textbf{BioBERTLSTM} & \textbf{RoBERTaGRU} & \textbf{RoBERTaLSTM} \\
    \midrule
    \textbf{CausalTimeBank}     & 0.636 & \textbf{0.645} (FTR) & 0.590 & 0.609 (BiLSTM) & 0.600 & 0.605 (FTR) \\
    \textbf{CauseNet-cause}     & 0.847 & 0.846 (FTR) & \textbf{0.848} & 0.843 (BiGRU) & 0.841 & 0.836 (BiGRU) \\
    \textbf{CauseNet-noncause}  & 0.833 & 0.833 (FTR) & \textbf{0.835} & 0.833 (FTR) & 0.828 & 0.822 (BiGRU) \\
    \textbf{FinCausal2020}      & 0.598 & \textbf{0.601} (FTR) & 0.596 & 0.594 (FTR) & 0.580 & 0.578 (FTR) \\
    \textbf{MedCaus}            & 0.773 & 0.770 (BiGRU) & 0.777 & 0.773 (BiGRU) & \textbf{0.778} & 0.773 (BiGRU) \\
    \textbf{SemEval 2018}       & 0.853 & 0.844 (BiGRU) & 0.850 & 0.838 (BiGRU) & \textbf{0.861} & 0.848 (BiGRU) \\
    \bottomrule
  \end{tabular}
\end{table*}

\section{Datasets} \label{sec_data}
We focus on model design and data augmentation for sentence-level causal relation extraction since most recent efforts in mining causal knowledge from the web sentence level \cite{moghimifar2020domain, heindorf2020causenet}. Unlike previous work, however, we distinguish sentences that express causality implicitly or explicitly and make sure to analyze both. The results from sentence-level explicit causal relation extraction could inform future work on multi-sentence causal relation and even document-level causal relation extraction.


Causal-TimeBank \cite{mirza2014annotating} consists of causal annotations of the TempEval-3 corpus \cite{uzzaman2013semeval}, which consists of news articles from the web. For uniformity across other datasets, we only consider sentence-level relations, although Causal-TimeBank also contains document-level relations.

CauseNet \cite{heindorf2020causenet} is a large graph of explicit causal relations and supporting sentences from ClueWeb12 and Wikipedia, with a high-precision subset that we utilized. For our purposes, we subsampled a collection of 5,000 sentences containing explicit markers such as "cause," "caused," and "causing" (CauseNet-cause). Another subsample from CauseNet includes 5,000 sentences without variants of the "cause" marker (CauseNet-noncause) but with other explicit causal markers like "leads to" and "due to." 

The FinCausal2020 dataset \cite{mariko2021financial} is a benchmark for the detection and extraction of causal relations in financial text. It was mined from financial news articles from various economics and finance websites. For our purposes, FinCausal was limited to relations contained in single sentences.

MedCaus \cite{moghimifar2020domain} is a dataset consisting of causal sentences mined from "medical articles" in  Wikipedia that matched certain seed patterns. While we found that many sentences in this dataset are medical or biological, some general sentences (E.g., ``The eastern water is saltier because of its proximity to Mediterranean Water") seem to be captured as well, so we have labeled it as a "General" domain dataset.

SemEval 2010 Task 8 \cite{hendrickx2019semeval} is a multi-way classification dataset. It has widely been used as a general domain benchmark for evaluating relation extraction tasks. Causal relation extraction literature has particularly focused on the Cause-Effect relations in this data which represent 12.4\% of the entire dataset. We use only the Cause-Effect relations for our analysis.

\section{Experiments on Model Architecture} \label{sec_model}
A first step towards open-domain, large-scale causal relation extraction is finding a model that works well across different datasets. We explore this through experiments on embeddings, recurrent units, and conditional random fields (CRFs), analyzing components used by high-performing models like SCITE \cite{li2021causality}. Since most models are developed on a single dataset, comparing them in the transfer setting can be misleading. For example, SCITE's \cite{li2021causality} annotation scheme is specific to SemEval, so some unifying simplifications should be made across datasets for fair comparison. \citet{yang2022survey} analyze recent model performance outside the transfer setting. Our experiments cover these models using combinations of tested components (e.g., SCITE is Flair-BiLSTM-CRF).

We are focused on extracting causal linkages on a sentence level, emphasizing intra-sentence relationships. \textcolor{black}{"Single Sentence-level"  described in this paper involves locating and identifying causal linkages in one sentence.} Our reason is that intra-sentences are label consistent across various datasets and our attempt to evaluate our model on multiple datasets justifies this focus. Furthermore, this allows us to take advantage of robust assessment measures and available annotated datasets to systematically improve our models before further expanding our scope to more complex inter-sentence causal relations, which involves identifying causal connections spanning multiple sentences and possibly entire documents.

Sentences for all datasets were tokenized at the word level \cite{akbik2019flair}, dependency parsed \cite{spacy2}, and each token was labeled as C (cause), E (effect), or O (other). Each dataset was split into 70\%-30\% train-test subsets. All models used an embedding layer, followed by a recurrent NN (RNN) layer, and either a CRF or linear layer for final prediction. Cross-entropy was used as the loss function. The RNN hidden size was fixed at 256 (empirically determined). 
Early stopping was used in training with a minimum of ten epochs. For sections 5 and 6, twelve training epochs were used as this was found to be consistently sufficient. An ADAM optimizer with a learning rate of 0.001 and a minibatch size of 16 was used.

\subsection{Embedding Input}
The trend in causal relation extraction models is to encode tokens as vectors using contextual word embeddings like BERT and Flair \cite{li2021causality}. However, an empirical analysis of the ideal embedding choice has not been conducted. We compare base BERT \cite{devlin2018bert}, BioBERT \cite{lee2020biobert}, and RoBERTa \cite{liu2019roberta} in the BERT category, and three Flair embeddings \cite{akbik2018coling} trained on general-purpose data, news data, and fine-tuned on CauseNet. Table \ref{embedding} shows that BERT, specifically BioBERT, performed the best overall. For each iteration in all the tables, each experiment was reproduced 10 times with different seed values. RoBERTa outperformed BioBERT for the SemEval 2018 dataset, and Base BERT outperformed BioBERT for the CausalTimeBank dataset. Additionally, a p-value analysis for 10 iterations of the MedCaus dataset showed no significant difference between BioBERT and RoBERTa embeddings, nor between BioBERT and Base BERT for the FinCausal2020 dataset. BioBERT's superior performance can be attributed to its training on a vast biomedical corpus, enabling it to capture scientific language nuances more effectively than its counterparts \cite{lee2020biobert}, however, BioBERT still excelled in FinCausal above its counterparts. 

\begin{table*}
  \caption{Macro-averaged F1 Scores for 10 iterations of Relation Extraction Models with and without CRF, and Hypothesis Testing Results}
  \label{crf}
  \centering
  \begin{tabular}{ccc|cc|cc}
    \toprule
    \textbf{Dataset} & \textbf{Base} & \textbf{Base+CRF}  & \textbf{BioBERT} & \textbf{BioBERT+CRF}  & \textbf{RoBERTa} & \textbf{RoBERTa+CRF} \\
    \midrule
    \textbf{CausalTimeBank}     & \textbf{0.636} & 0.636  (FTR) & 0.590 & 0.597 (FTR) & 0.600 & 0.591 (NoCRF) \\
    \textbf{CauseNet-cause}     & 0.847 & 0.847 (FTR) & 0.848 & \textbf{0.849} (FTR) & 0.841 & 0.842 (FTR) \\
    \textbf{CauseNet-noncause}  & 0.833 & 0.836 (CRF) & 0.835 & \textbf{0.837} (FTR) & 0.828 & 0.829 (FTR) \\
    \textbf{FinCausal2020}      & \textbf{0.598} & 0.591 (FTR) & 0.596 & 0.587 (FTR) & 0.580 & 0.578 (FTR) \\
    \textbf{MedCaus}            & 0.773 & 0.771 (FTR) & 0.777 & 0.776 (FTR) & \textbf{0.778} & 0.777 (FTR) \\
    \textbf{SemEval 2018}       & 0.853 & 0.850 (FTR) & 0.850 & 0.849 (FTR) & 0.861 & \textbf{0.862} (FTR) \\
    \bottomrule
  \end{tabular}
\end{table*}

\subsection{Recurrent Unit}
Many successful causal relation extraction models rely on LSTM \cite{hochreiter1997long} recurrent units, particularly BiLSTMs \cite{li2021causality}, to encode sequential information and parse causal relations in both text directions. Despite the common use of BiLSTMs, the optimal recurrent unit, especially compared to BiGRUs \cite{cho2014properties}, is underexplored. Our experiment aimed to address this by keeping all hyperparameters constant except for the recurrent unit. The results, detailed in Table \ref{recurrent} indicate that BiGRUs outperform BiLSTMs. The analysis was done at a 5\% significance level for 10 iterations. FTR means Fail to Reject, which suggests that while the choice of the recurrent unit slightly impacts overall performance, BiGRUs' efficiency and reduced complexity, with fewer parameters and gates, offer subtle advantages in processing causal relations. Our analysis, supported by p-value assessments across various datasets, reveals no significant performance difference between the embeddings of BioBERT, RoBERTa, and Base BERT in most cases. These findings align with the literature suggesting that advanced contextual embeddings, such as those from BERT and Flair, proficiently capture long-term dependencies, potentially diminishing the impact of additional LSTM gates. This empirical evidence challenges the assumption of a one-size-fits-all approach to recurrent units in causal relation extraction, highlighting the importance of embedding choice and its compatibility with the model architecture.

\subsection{Conditional Random Fields}
Conditional random fields (CRF) \cite{lafferty2001conditional} are probabilistic graphical models that can improve segmentation when integrated with neural sequence classifiers \cite{zheng2015conditional}. While some causal relation extraction models have successfully used CRFs \cite{li2021causality}, others have opted not to \cite{moghimifar2020domain}. Some evidence suggests that a well-encoded NN may not need a CRF layer \cite{lample2016neural}. We conducted experiments to test the utility of CRFs across different causal relation extraction datasets.

A comparison of models with and without CRF is shown in Table \ref{crf}, with its p-value analysis. FTR means Fail to Reject, meaning that there was no significant difference between the CRF and No-CRF experiments.  From the table, the p-value analysis indicated no significant difference, especially for BioBERT embedding, across all datasets. Only in the case of CauseNet-noncause did we see a significant difference at 5\%, favoring CRFs when using BaseBERT embedding, but overall, BioBERT embedding still performed better. Given this and the potential for over-fitting in open-domain settings due to varying entity lengths, we excluded CRF from further analysis.

\section{Comparing Transfer Performance}
\subsection{Phrasal Loss: $F1_{phrase}$}
\begin{figure*}
\includegraphics[width=1\textwidth]{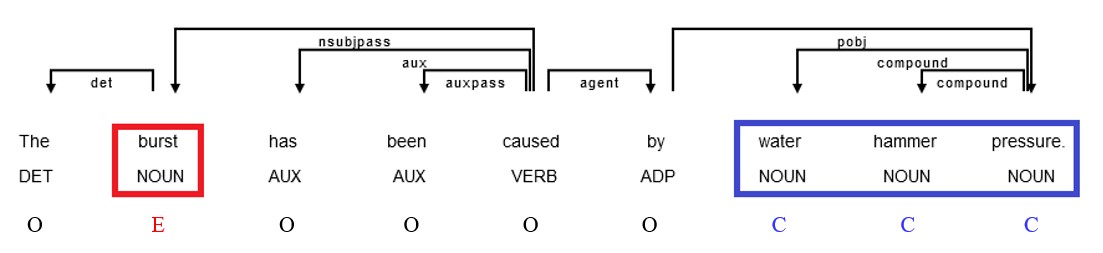}
\caption{
The $F1_{phrase}$ metric measures a models ability to locate these noun phrases. If a candidate model were to predict "hammer" as the only cause token, the entire phrase "water hammer pressure" would be counted as entirely correct because the phrase can completely be recovered from "hammer" using the dependency tree.
} 
\label{f1}
\end{figure*}

A roadblock in causal transfer is a lack of annotation consistency across datasets. For example, SemEval consists of mostly single-word cause/effect annotations, with very few multi-word causes and effects. MedCaus has lengthier cause/effect phrases that span more words on average. On the extreme end, FinCausal annotations mark most tokens in a sentence as C or E, with very few O. This makes transfer difficult, as the objective varies across target data. Naturally, we would expect a model trained on single-word annotations to only identify single words in target data. Since root nouns can be recovered with dependency parsing, penalizing this behavior is excessively harsh. To remedy this, we introduce $F1_{phrase}$, which is defined as follows.

$$F1_{phrase}(\mathbf{\tilde{y}}, \mathbf{y}) = F1(\lambda(\mathbf{\tilde{y}}), \lambda(\mathbf{y})) $$
$$F1_{phrase}(\mathbf{\tilde{y}}, \mathbf{y}) = \frac{2}{\mathrm{precision}(\lambda(\mathbf{\tilde{y}}), \lambda(\mathbf{y}))^{-1}  +  \mathrm{recall}(\lambda(\mathbf{\tilde{y}}), \lambda(\mathbf{y}))^{-1}}$$
Where $\mathbf{y} = (y_1, y_2, ..., y_t)$ is a target sequence of token labels and  $\mathbf{\tilde{y}} = (\tilde{y}_1, \tilde{y}_2, ..., \tilde{y}_t)$ is the predicted set of token labels. The function $\lambda$ is an element-wise function with oracle knowledge of cause effect phrases defined as follows:

\begin{equation}
\lambda(z_i) =
    \begin{cases}
        z_i                  & \text{if } z_i \text{ \textcolor{black}{is not in a cause/effect phrase}}\\
        \text{parent}(z_i) & \text{if } z_i \text{ \textcolor{black}{is in a cause/effect phrase}}
    \end{cases}
\end{equation}

In our case, the $parent(z_i)$ function is based on dependency parsing, considering the "compound" or "amod" dependency relations \cite{spacy2}. This is, in essence, the F1 score modified so that if any token in the cause-effect phrase is identified correctly, then that entire phrase is counted as correct. This is acceptable because after relations are mined, root nouns can be recovered from noun phrases (and vice versa) via dependency parsing, as illustrated in Figure \ref{f1}.

To illustrate the utility of $F1_{phrase}$ for evaluating transfer performance, we can look at the specific case of transferring from CauseNet to SemEval. CauseNet has a longer mean entity length (cause: 1.6 tokens, effect 1.5 tokens), than SemEval (cause: 1.06 tokens, effect: 1.02 tokens). Furthermore, the part-of-speech distributions vary between the two datasets with respect to token labels. CauseNet has disproportionately many more adjectives than SemEval 
Training a RoBERTa-based sequence classifier on CauseNet and evaluating on SemEval yielded an unimpressive F1 score of 67.4\% (precision 0.632, recall 0.721). However, this would be an unfair evaluation. In the case of the sentence in Figure \ref{f1}, the model would likely predict ``pressure" as the only cause token, but such a prediction is in essence correct. The $F1_{phrase}$ gives a more honest score of $F1_{phrase} = 83.4\%$, (precision 0.978, recall 0.727). 

\begin{table}[h]
\centering
\caption{Comparison of F1 Phrase Score and F1 Score across Datasets}
\label{Fphrase}
\begin{tabular}{lccc}
    \toprule
    \textbf{Dataset} & \textbf{F1 Phrase Score} & \textbf{F1 Score} & \textbf{\% Change} \\
    \midrule
    CausalTimeBank    & 0.719 & 0.590 & 21.86\% \\
    CauseNet\_cause   & 0.874 & 0.848 & 3.07\% \\
    CauseNet\_noncause& 0.877 & 0.835 & 5.03\% \\
    FinCausal2020     & 0.844 & 0.596 & 41.61\% \\
    MedCaus           & 0.985 & 0.777 & 26.77\% \\
    SemEval\_2018     & 0.921 & 0.850 & 8.35\% \\
    \bottomrule
\end{tabular}
\end{table}

\textcolor{black}{In Table \ref{Fphrase}, the F1 score is compared to the F1-Phrase. CausalTimeBank shows a 21.86\% improvement, highlighting the challenge of capturing implicit causality within temporal and event-driven texts, where understanding context is key. In contrast, CauseNetcause exhibits a smaller 3.07\% change, reflecting its focus on explicit causal relationships that are more straightforward to identify. CauseNetnoncause, with a 5.03\% change, reveals the difficulty in distinguishing causal from non-causal associations, nonetheless, made up of simpler sentences. FinCausal2020, with a significant 41.61\% change, underscores the complexity of identifying causality in financial texts, where implicit relationships are often embedded in technical language. MedCaus, showing a 26.77\% change, shows the mix of explicit and implicit causality in medical texts, where complex terms and conditional statements are common. Lastly, SemEval 2018 demonstrated an 8.35\% change, reflecting the diverse nature of its text types, which require a nuanced understanding of various causal expressions.}

\subsection{Pairwise transfer and Combined Data Transfer}
The pairwise transfer was conducted by training a BioBERT-BiGRU model on one dataset and then evaluating it on another dataset's test set, creating the matrix of transfer performance values shown in Table \ref{pairwise}. This matrix averages 10 iterations. Baseline model performance was also assessed, as shown on the main diagonal of Table \ref{pairwise}. $F1_{phrase}$ was used for evaluations, calculated using dependency trees generated from spaCy \cite{spacy2}. Results in Table \ref{pairwise} show that transfer from MedCaus outperforms training on the test set's own data for the $CauseNet_{\text{cause}}$ and CausalTimeBank datasets. The $CauseNet_{\text{non-cause}}$ dataset evaluated by p-value shows no significant difference. Notably, MedCaus, the largest dataset studied, has longer cause/effect phrase annotations. We investigate the impact of transfer data size in the following section. Longer annotations seem advantageous when using $F1_{phrase}$, which focuses on noun phrase localization. Thus, a good approach for general-purpose causal relation extraction models could be training on datasets with longer annotations. If only the root nouns are needed after extraction, they can be recovered with dependency parsing.

\begin{table*}
\centering
\resizebox{\textwidth}{!}{
\begin{tabular}{l|lllllll}

\hline
\textbf{Dataset}          & \textbf{CausalTimeBank} & \textbf{CauseNet\_cause} & \textbf{CauseNet\_noncause} & \textbf{FinCausal2020} & \textbf{MedCaus} & \textbf{SemEval\_2018} \\ 
\hline
\textbf{CausalTimeBank}   & 0.719 & 0.445 & 0.509 & 0.374 & 0.534 & 0.569 \\
\textbf{CauseNet\_cause}  & 0.446 & 0.874 & 0.790 & 0.244 & 0.666 & 0.809 \\
\textbf{CauseNet\_noncause} & 0.507 & 0.802 & \textbf{0.877} & 0.375 & 0.693 & 0.796 \\
\textbf{FinCausal2020}     & 0.809 & 0.767 & 0.771 & \textbf{0.844} & 0.813 & 0.778 \\
\textbf{MedCaus}       & \textbf{0.847} & \textbf{0.881} & 0.874 & 0.755 & \textbf{0.985} & 0.881 \\
\textbf{SemEval\_2018}   & 0.498 & 0.731 & 0.703 & 0.350 & 0.624 & \textbf{0.921} \\ \hline

\end{tabular}
}
\caption{
Matrix of $F1_{phrase}$ scores for transferred relation extraction models, with rows correspond to the training data and columns correspond to the test set used. For the main diagonal, the training data is split from the complete dataset. Otherwise, the entirety of the training data is used. The F1-phrase scores shown in bold correspond to transferred datasets that outperformed the test sets own training data.
}

\label{pairwise}
\end{table*}

We should also examine if including multiple datasets of different domains improves or detracts from transfer performance. For example, if SemEval was the target dataset, then SemEval's training data and the entirety of all other datasets were used for training as shown in Table \ref{pairwise},  

In all cases, including additional datasets significantly improved performance above the baseline. For example, the CauseNetcause dataset saw an increase in $F1_{phrase}$ from 0.874 to 0.882, a modest improvement of 0.92\%. Similarly, MedCaus improved from 0.985 to 0.995, marking a 1.01\% increase. CauseNetnoncause also benefited, with an improvement from 0.877 to 0.886, translating to a 1.03\% increase. Notably, CausalTimeBank and FinCausal2020 showed substantial improvements, with $F1_{phrase}$ scores increasing from 0.719 to 0.747 (3.89\%) and from 0.844 to 0.883 (4.62\%), respectively. This is particularly significant given that both datasets have relatively longer annotations than most of the data used for training. Even though CausalTimeBank started as the smallest dataset, it achieved the second most significant improvement. SemEval2018 also saw a boost in performance, with the $F1_{phrase}$ score increasing from 0.921 to 0.929, a 0.87\% improvement. These results clearly indicate that regardless of the dataset's lexical composition, domain, or size, incorporating a more varied training dataset leads to better causal relation extraction.

\section{Qualitative Analysis}
The larger size of the MedCaus dataset, which contains 8,682 sentences, maybe a contributing factor to its effective transfer. To explore this hypothesis, training was conducted using progressively larger subsets of MedCaus, starting with 10\% and increasing incrementally by 10\% up to the full dataset. The outcomes of these trials suggested that expanding the training subset beyond 40\% of the MedCaus dataset does not significantly enhance the $F1_{phrase}$ score for the majority of datasets, with the notable exception of the CausalTime Bank dataset. Additionally, there appears to be no negative impact from using larger amounts of data; however, it seems to be a superfluous and potentially inefficient use of computational resources, considering the plateau in average performance observed when training with 40\% to 60\% of the data. This indicates that simply increasing the size of the dataset used for transfer does not necessarily correlate with better model performance.


Another consideration for causal transfer is the composition of implicitly and explicitly causal sentences in the training data. \cite{chen2023deep} While explicit relations like those in CauseNet are more readily available, a relation extraction model trained on explicit relations alone might merely learn the syntactic templates that generated CauseNet, and therefore fail to generalize to implicit relations. To test this, the BioBERT-BiGRU model was trained on implicit and explicit subsets of MedCaus, at accumulating larger sizes, then tested on SemEval, Causal-TimeBank, and CauseNet. 
Training with only explicit sentences from MedCaus leads to better performance on the CauseNet datasets since these are all based on explicit syntactic templates. For SemEval, equal-sized subsets of explicit and implicit sentences seemed to provide equal transfer performance. For CausalTimeBank, training with implicit sentences was better. Since CausalTimeBank is over half implicit (54.7\% over SemEval's 34\%). Thus, the implicit/explicit composition of the training data should reflect the composition of the target data in the transfer setting.

\section{Discussion}

Our experiments show that a BioBERT-BiGRU sequence tagger is a strong candidate for large-scale causal relation extraction in the open domain. BioBERT's pretraining with dynamic masked language modeling seems to enhance model performance, suggesting that dynamic masking for causal phrases is worth investigating. Causality-specific pretraining/fine-tuning, as demonstrated by \citet{khetan2022causal} and \citet{li2021causality}, is further evidenced by our results. Future research could explore pretraining for open-domain relation extraction.

While our current study focuses on intra-sentence causal relation extraction, this approach allows us to fine-tune our models and metrics in a more manageable context. Addressing intra-sentence relations first ensures that our methods are well-grounded and effective within simpler boundaries. This step-by-step approach is crucial for developing the sophisticated techniques required for inter-sentence causal relation extraction, which will also be the focus of future research.

The $F1_{phrase}$ metric reveals that combining disparate data from varying annotation schemes and domains can significantly improve transfer performance in causal relation extraction. No single dataset is likely to be the key to large-scale causal relation extraction, but combining diverse data has proven beneficial. Future work could benefit from combining even more causal datasets than we used here. Our results from section 6 show that data size matters up to a point, but increasing training data beyond a few thousand sentences does not always improve transfer performance. Instead, attention should be given to the composition of implicit and explicit sentences in the dataset.

\section{Limitations}
\textcolor{black}{This study did not consider advanced large language models like GPT, Mistral-7B, and Llama. Although these recent models may have the potential to offer enhanced performance, our study is confined to using BERT and Flair embedding, because we considered the optimal balance between computational efficiency and robust performance in domain-specific tasks. Moreover, BERT and Flair's embedding are well-documented and widely integrated into existing NLP pipelines, making them a practical and reliable choice for investigating causality in this context.} 

\textcolor{black}{Also, this study was limited to sentence-level causal relation extraction, when inter-sentence and document-level relations offer a rich source of causal information for open-domain applications. While some of the datasets used, such as FinCausal and CausalTimeBank, do contain inter-sentence relations, we limited our analysis to sentence-level relations for compatibility with other datasets.}

Although this study includes a range of datasets with varying domains and annotation styles, the selection may still not fully capture the diversity of causal relations present in real-world texts. Additional datasets from different domains and with different annotation styles could further validate the findings. The models and methods were primarily tested on specific benchmark datasets. While efforts were made to ensure transferability, the generalizability to other, non-tested datasets and domains cannot be assumed and therefore, remains uncertain.

While our goal was to use a simple model like BioBERT-BiGRU which showed promising results, this study did not explore more advanced architectures that might offer improved performance or interpretability. It is possible in the future to explore the trade-offs between model complexity and performance in more detail.

\section{Conclusion}

We conducted an empirical study of model design and data augmentation strategies for building causal relation extraction models that transfer effectively. We showed BioBERT embeddings paired with a relatively simple BiGRU, is a strong performing base causal relation extraction model across varying datasets. To better facilitate data transfer experiments, we introduce $F1_{phrase}$, a variant of the F1 measure that prioritizes noun phrase localization. This revealed that longer annotations transfer well, cross-domain data augmentation is consistently beneficial, and implicit/explicit dataset composition trumps dataset size in terms of transfer performance.

\section {Ethical Considerations}
\begin{enumerate}
\item {All data used in this experiment is open-source and most are licensed by Creative Commons. The others had their individual licensing}
\item{Non-sensitive data was used during this research and as such, no privacy violations were made.}
\item{There was clear documentation of methodologies and algorithms to enable transparency, reproducibility, and accountability.}
\item {For each of the datasets, we followed their individual terms of usage}
\end{enumerate}


\bibliographystyle{ACM-Reference-Format}
\bibliography{ref}

\section {License}
\begin{enumerate}
\item{The Causal-TimeBank corpus is provided under the Creative Commons Attribution-ShareAlike 4.0 International (CC BY-SA 4.0) license. \href{https://paperswithcode.com/datasets/license}{Click here for more details} }
\item{CauseNet is available for non-commercial academic use under the Creative Commons Attribution-NonCommercial 4.0 International (CC BY-NC 4.0) license. \href{For detailed terms, visit https://opensource.org/license/MIT}{Click here for detailed terms} }

\item{The FinCausal2020 dataset is provided under the Creative Commons Attribution-NonCommercial 4.0 International (CC BY-NC 4.0) license. \href{For detailed terms, visit  https://www.thefinancialguide.com/licenses}{Click here for detailed terms} }
\item{
The MedCaus dataset, derived from Wikipedia and medical articles, is available under the Creative Commons Attribution-ShareAlike 4.0 International (CC BY-SA 4.0) license. 
}
\item {The SemEval 2010 Task 8 dataset is available for research purposes under the Creative Commons Attribution-ShareAlike 4.0 International (CC BY-SA 4.0) license.
}
\end{enumerate}


\appendix
\section{Appendix} \label{sec:appendix}

\begin{figure*}
\includegraphics[width=1\textwidth]{ 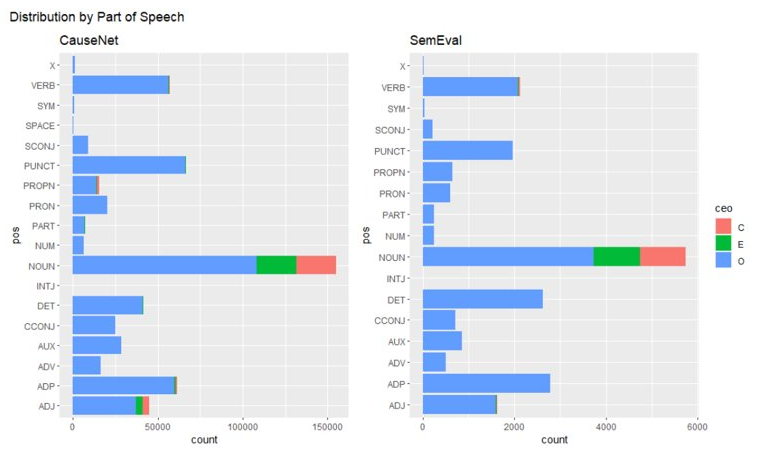}
\caption{
Part of speech distributions for CauseNet and SemEval after dependency parsing. The difference in annotation schemes between CauseNet and SemEval is clear, as disproportionately more adjectives are fall under ``C" (cause) and ``E" (effect) labels in CauseNet. The $F1_{phrase}$ metric accounts for this discrepancy. 
} \label{postdist}
\end{figure*}

\begin{table*}
\centering
\begin{tabular}{llllll}
\hline
\textbf{Dataset} & \textbf{Sentences} & \textbf{Implicit} & \textbf{Domain} & \textbf{Mean tokens per C/E} & \textbf{Reference}\\
\hline
MedCaus & 8682 & 17\% & General & 8.41 / 7.68 & \citealp{moghimifar2020domain}\\
CauseNet-noncause & 5000 & 0\% & General & 1.61 / 1.5 & \citealp{heindorf2020causenet}\\
CauseNet-cause & 5000 & 0\% & General & 1.53 / 1.46 & \citealp{heindorf2020causenet}\\
SemEval 2010 & 1003 & 34\% & General & 1.06 / 1.02 & \citealp{hendrickx2019semeval}\\
CausalTimeBank & 298 & 54.7\% & News & 1 / 0.99 & \citealp{mirza2014annotating}\\
FinCausal2020 & 1719 & 78.7\% & Financial & 23.72 / 10.26 & \citealp{mariko2021financial}\\
\hline
\end{tabular}
\caption{
Sentence level causal relation extraction datasets used in this paper. Percentage of implicit sentences in each dataset is shown. Implicit is defined as a sentence that does not fit the syntactic patterns mined by CauseNet. "Mean tokens per C/E" gives the average length, in tokens, of cause and effect annotations across the dataset.
}
\label{datasets}
\end{table*}

\begin{table*}
    \centering
    \begin{tabular}{|l|l|l|l|l|l|}
    \hline
        Model Type & Train F1 & Test F1 &  Epochs & Split & Params \\
    \hline
Fine Tuned Flair-BiGRU & 0.945 & 0.811 & 8 & 2000/1000 & hidden = 32 \\
Fine Tuned Flair-BiGRU & 0.982 & 0.808 & 8 & 2000/1000 & hidden = 64 \\
Fine Tuned Flair-BiGRU & 0.991 & 0.819 & 8 & 2000/1000 & hidden = 128 \\
Fine Tuned Flair-BiGRU & 0.995 & 0.820 & 8 & 2000/1000 & hidden = 256 \\
Fine Tuned Flair-BiGRU & 0.997 & 0.811 & 8 & 2000/1000 & hidden = 512 \\
Fine Tuned Flair-BiGRU & 0.998 & 0.815 & 8 & 2000/1000 & hidden = 1024 \\
    \hline
    \end{tabular}
    
\caption{
Performance of a Flair-BiGRU causal relation extractor across varying hidden dimension. A hidden dimension of 256 leads to the best test F1. 
}
\label{app_hidden}
\end{table*}

\begin{table*}
    \centering
    \begin{tabular}{|l|l|l|l|l|l|}
    \hline
        Model Type & Train F1 & Test F1 & SemEval F1 & Epochs & Split \\
    \hline

BioBERT-BiGRU	& 0.995	& 0.724 &	0.676	 & 12 &	200/100/1003	\\
BioBERT-BiGRU	& 0.999	& 0.632 &	0.695	 & 12 &	2000/1000/1003	\\
BioBERT-BiGRU	& 0.987	& 0.873 &	0.69	 & 12 &	20000/1000/1003	\\
    \hline
    \end{tabular}
    
\caption{
This table shows performance of a BioBERT-BiGRU model trained on subsamples of CauseNet that accumulate in size, with transfer performance on SemEval. Increasing the training data by an order of magnitude beyond 2,000 does not improve generalization performance. The information in the training data that can be used meaningfully on the transfer target is learned within a few thousand samples.
}
\label{app_trainsize}
\end{table*}

\end{document}